\pdfoutput=1

\documentclass[11pt]{article}

\usepackage[]{EMNLP2023}

\usepackage{times}
\usepackage{latexsym}
\usepackage{amsmath}
\usepackage{amssymb}
\usepackage{fixmath}
\usepackage{graphicx}
\usepackage{float}
\usepackage{xspace}
\usepackage[normalem]{ulem}
\usepackage{multirow}
\usepackage{multicol}
\usepackage{caption}
\usepackage{subcaption}

\usepackage[T1]{fontenc}

\usepackage[utf8]{inputenc}

\usepackage{microtype}

\usepackage{inconsolata}
\usepackage{enumitem}

\usepackage{booktabs}
\usepackage{diagbox}
\newcommand{\our}{\textsc{XAI-Class}\xspace}

\title{\our: Explanation-Enhanced Text Classification with Extremely Weak Supervision}

\author{Daniel Hajialigol, Hanwen Liu, Xuan Wang \\
  Virginia Tech, VA, USA \\
  \texttt{\{danielhajialigol,liuhwen,xuanw\}@vt.edu} \\
  }

\begin{document}
\maketitle

\begin{abstract}
Text classification aims to effectively categorize documents into pre-defined categories. Traditional methods for text classification often rely on large amounts of manually annotated training data, making the process time-consuming and labor-intensive. To address this issue, recent studies have focused on weakly-supervised and extremely weakly-supervised settings, which require minimal or no human annotation, respectively.
In previous methods of weakly supervised text classification, pseudo-training data is generated by assigning pseudo-labels to documents based on their alignment (e.g., keyword matching) with specific classes. However, these methods ignore the importance of incorporating the explanations of the generated pseudo-labels, or \textit{saliency} of individual words, as additional guidance during the text classification training process. 
To address this limitation, we propose \our, a novel explanation-enhanced extremely weakly-supervised text classification method that incorporates word saliency prediction as an auxiliary task. \our begins by employing a multi-round question-answering process to generate pseudo-training data that promotes the mutual enhancement of class labels and corresponding explanation word generation. This pseudo-training data is then used to train a multi-task framework that simultaneously learns both text classification and word saliency prediction. Extensive experiments on several weakly-supervised text classification datasets show that \our outperforms other weakly-supervised text classification methods significantly. Moreover, experiments demonstrate that \our enhances both model performance and explainability.
\end{abstract}

\section{Introduction}
\begin{figure}
    \centering
    \includegraphics[width=75mm]{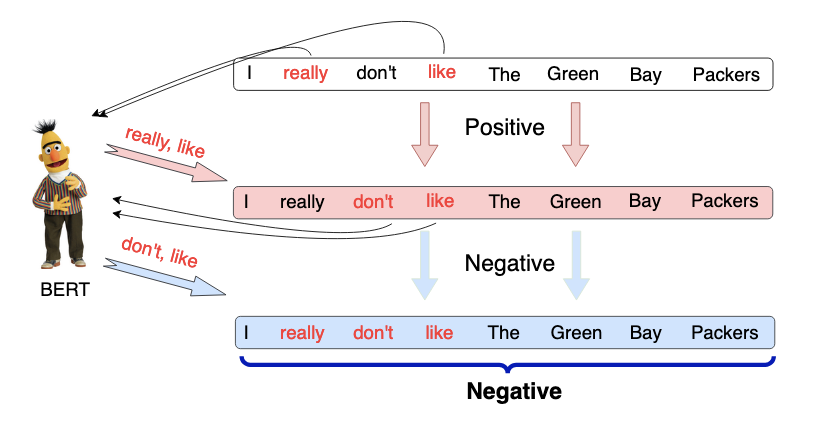}
    \caption{Previous weakly-supervised text classification methods do not model salient words, potentially leading to uncertain predictions. On the other hand, \our generates pseudo-text classification and pseudo-saliency labels by querying two pre-trained language models (PLMs) and updating pseudo-saliency labels by using previously generated pseudo-text classification labels and vice-versa.}
    \label{fig:enter-label}
\end{figure}

Text classification is a fundamental task in natural language processing (NLP), aiming to effectively categorize documents (e.g., news reports) into pre-defined categories (e.g., politics, sports, and business). It has various downstream applications such as information extraction \cite{zhang-event-2022}, sentiment analysis \cite{tang-etal-2015-document}, and question answering \cite{rajpurkar-etal-2016-squad}. 

Traditional methods for text classification \cite{yang-etal-2016-hierarchical, DBLP:journals/corr/abs-1906-08237,DBLP:journals/corr/ZhangZL15} often rely on large amounts of manually annotated training data, making the process time-consuming and labor-intensive. To address this issue, recent studies have focused on weakly-supervised \cite{dataless,10.5555/2892753.2892772,10.5555/1625275.1625535,badene-etal-2019-data,ratner2017data,westclass,conwea,10.1145/336597.336644,10.1145/3397271.3401121, 8594978} and extremely weakly-supervised \cite{lotclass,conwea,xclass,wddc,zhang-etal-2021-weakly} settings, which require minimal or no human annotation, respectively. 
In this study, we focus on the extremely weakly-supervised setting that utilizes only the class names as supervision. Importantly, we do not assume that the class names need to have appeared in the input documents. 

Previous methods for extremely weakly-supervised text classification usually start with finding initial keywords for each class to construct a keyword vocabulary. This vocabulary is then employed to assign pseudo-labels to documents, followed by training the model using traditional supervised learning techniques. For example, LOT-Class \cite{lotclass} leverages a pre-trained masked language model to predict keywords that can replace label words. However, this method assumes that the class names must appear in the input document, which may not be feasible in many real-world scenarios. Recent advancements have relaxed this constraint and do not assume that the class names need to have appeared in the input documents. For example, X-Class \cite{xclass} obtains the word and document representations and employs clustering methods for keyword grouping and label assignment, while WDDC \cite{wddc} applies cloze-style prompting to identify keywords and assigns pseudo-labels based on the representation similarity between the keywords and the documents.
However, previous methods ignore the importance of incorporating the explanations of the generated pseudo-labels, or \textit{saliency} \cite{simonyan2014deep} of individual words, as additional guidance during the text classification training process (Figure \ref{fig:enter-label}).  This oversight has limited the potential of these methods to fully exploit the valuable insights provided by explanations and word saliency that can greatly enhance the effectiveness and explainability of the text classification methods.

To address this limitation, we propose \our, a novel explanation-enhanced extremely weakly-supervised text classification method that incorporates word saliency prediction as an auxiliary task. \our begins by employing a multi-round question-answering process to generate pseudo-training data that promotes the mutual enhancement of class labels and corresponding explanation word generation. Specifically, we first leverage a pre-trained multi-choice question-answering model \cite{flant5} to query the predicted class labels for given documents. Using the predicted class labels as input, we then query a pre-trained extractive question-answering model \cite{bert} to identify the tokens in the document that were most influential in predicting the class labels. This iterative process continues until the predictions remain consistent, indicating high confidence in both the predicted class labels and the saliency words. The resulting pseudo-training data incorporates both the class labels and the associated explanation words. This pseudo-training data is then used to train a multi-task framework that simultaneously learns both text classification and word saliency prediction. By jointly optimizing both tasks, the model can effectively enhance both the performance and explainability of the text classification model.
Our contributions are summarized as follows:
\begin{itemize}[leftmargin=*]
    \item We propose \our, a novel extremely weakly-supervised text classification method that leverages multiple-round question answering to promote mutual enhancement between text classification and word saliency prediction pseudo-training data generation.
    \item We propose a novel explanation-enhanced text classification method that trains a multi-task framework to simultaneously learn both text classification and word saliency prediction.
    \item Experiments on several datasets demonstrate the superiority of \our over previous weakly-supervised text classification methods for both performance and explainability.
\end{itemize}
We will open-source our code and results as a baseline to facilitate future studies.


\section{Related Work}
\subsection{Text Classification Methods}  
Traditional methods for text classification \cite{yang-etal-2016-hierarchical, DBLP:journals/corr/abs-1906-08237,DBLP:journals/corr/ZhangZL15} often rely on large amounts of manually annotated training data, making the process time-consuming and labor-intensive. To address this issue, recent work has been proposed for text classification with minimal human annotation.

\paragraph{Weakly-Supervised Text Classification} To address the above issue of manual annotation, recent studies have focused on the weakly-supervised setting that requires minimal human annotation. For example, Snowball \cite{10.1145/336597.336644}  combines pattern-based and distant supervision techniques to extract relations. It uses patterns based on syntactic dependencies and entity mentions to identify potential relations in sentences. However, this pattern-based approach may struggle with complex relations involving multiple entities or deeper semantic understanding. Dataless \cite{dataless} proposes a classification method using semantic representation. It leverages external knowledge sources to capture the semantic information in the text. However, the limitation is its dependence on the availability and quality of external knowledge sources. Doc2cube \cite{8594978} clusters similar documents and assigns them to text cubes. It leverages the inherent structure and patterns within the collection for guidance. However, the effectiveness of Doc2Cube depends on the quality of document similarity measures used for clustering. Inaccurate or inadequate similarity metrics can impact document allocation accuracy.

\paragraph{Extremely Weakly-Supervised Text Classification}
Compared with weakly-supervised text classification, extremely weakly supervised text classification goes a step further by using even weaker supervision or no labeled data during training. For example, LOTClass \cite{lotclass} consists of three steps: substituting label names to enable the model to understand the meaning of each label, identifying category-relevant words for word-level classification, and finally conducting generalized self-training. Conwea \cite{conwea} utilizes contextualized word representations generated by PLMs to capture the rich semantic information of words in context for label assignment. XClass \cite{xclass} expands label words and generates document representations based on BERT \cite{bert} for clustering and the best documents are selected to train the classifier. WDDC \cite{wddc} uses cloze-style completion to generate summary text words, which serve as supervised signals for training the document classifier. However, these methods all have high requirements for the frequency of occurrence of labels and their closely related words in the text. 
ClassKG \cite{zhang-etal-2021-weakly} constructs a keyword graph by extracting important keywords from the documents, which serves as a representation of the document collection. Then ClassKG utilizes the connectivity and similarity of keywords in the graph to train the model. However, the efficiency and scalability of the method can be a concern when dealing with large-scale datasets. 

\subsection{Explainable Text Classification}
Explainable text classification methods can be decomposed into two categories: post-hoc explainability and intrinsic explainability. 

\paragraph{Post-hoc Explainability} Post-hoc explainability explain inputs \textit{after} a model has already been trained. This category consists of perturbation methods, such as LIME \cite{lime}, which learns an interpretable model of points in the neighborhood of a given input. Post-hoc explainability techniques can also be categorized by backpropagation-based methods. For example, \citeauthor{simonyan2014deep} attempt to explain instances by introducing the concept of saliency maps, which calculate gradients of inputs with respect to the inputs' features. \citeauthor{inputx} extends this idea by computing the partial derivatives of the prediction with respect to the input and multiplies them with the input \cite{ancona2017towards}.

\paragraph{Intrinsic Explainability}
In contrast to post-hoc explainability, intrinsic explainability methods attempt to create models that offer explanations. This has been accomplished through a handful of measures, one of which being constraining features \cite{freitas2014comprehensible} to be sparse and by measuring feature sensitivity \cite{simonyan2014deep}. \our aligns with this class of explainable text classification, as we generate and inject saliency information in our framework directly.

\section{Methodology}
\begin{figure*}
    \centering
    \includegraphics[width=160mm]{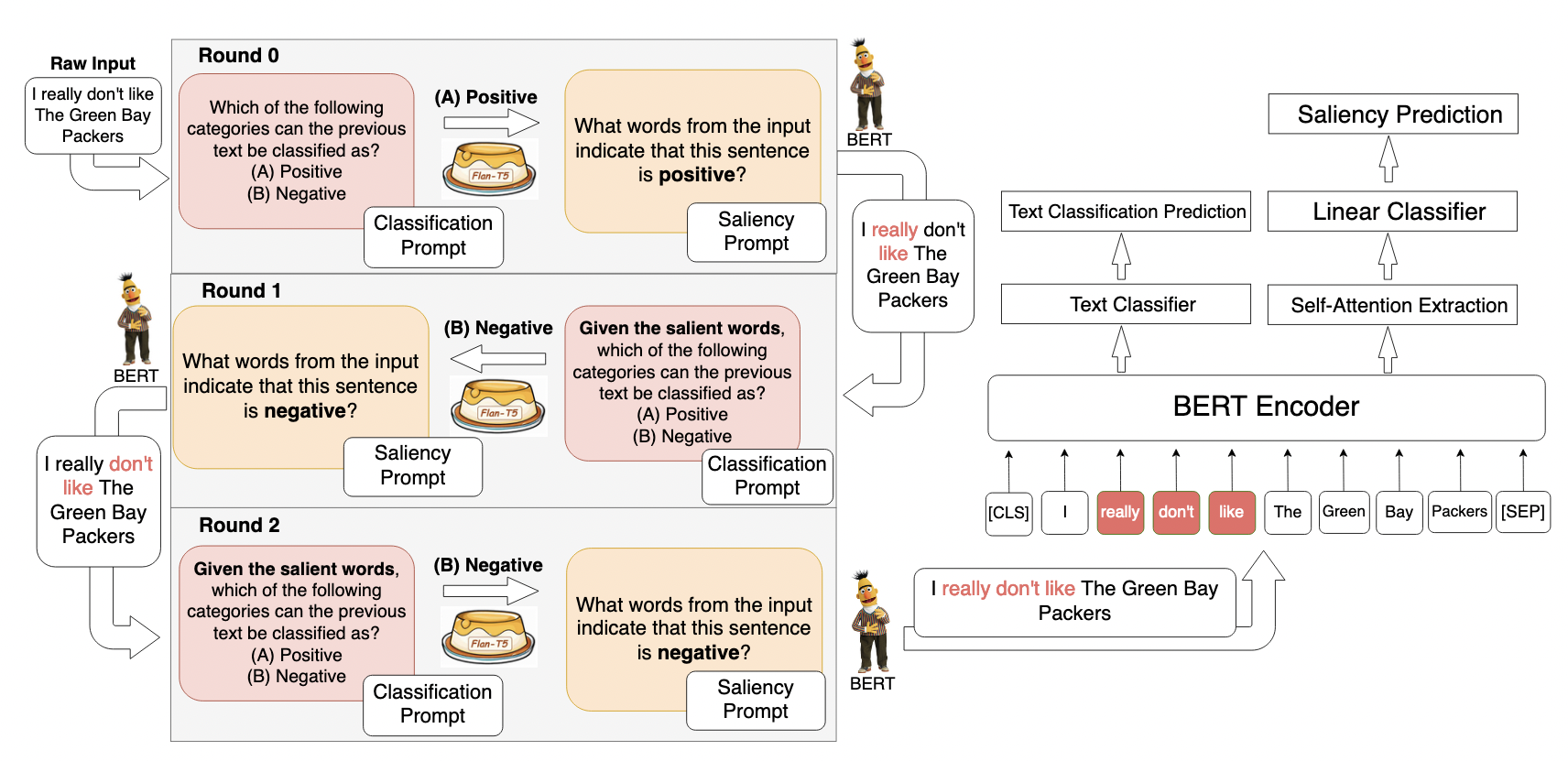}
    \caption{\our architecture. (Left) Given an input document $\mathcal{D}$ ("I really don't like The Green Bay packers"), we first query the class prediction from a PLMs $\mathcal{T}\tau$ (\textsc{FLAN-T5} \cite{flant5} in this figure) and then query the indicative words (highlighted in red) from another PLMs $\mathcal{T}^{E}$ (BERT \cite{bert} in this figure), forming our initial setup. We introduce the notion of a \textit{round}, where we once again query $\mathcal{T}\tau$ using the queried indicative words and use this more confident prediction to query the salient words from $\mathcal{T}^{E}$ once more. We repeat this operation until a variable number of rounds. (Right) We then tokenize $\mathcal{D}$ and feed this along with the salient tokens into our BERT-based \cite{bert} multi-task learning model, learning to predict both text classification and saliency labels using the contextualized representations.}
    \label{fig:main}
\end{figure*}

We propose \our, an explanation-enhanced extremely weakly-supervised text classification method. The \our framework (Figure \ref{fig:main}) consists of two major steps: (1) iterative pseudo-label generation, and (2) explainable multi-task learning. In this section, we describe the \our framework in detail.

\subsection{Preliminaries}
\paragraph{Problem Formulation}
Our framework operates under the extremely weakly supervised text classification scenario, whose goal is to predict the correct class of a document with only its contents and the possible classes it could be categorized into. Mathematically, we represent a corpus as $\mathcal{X}$ which contains documents $\mathcal{D} = \{t_{i}| \forall i \in [1, |\mathcal{D}|]\}$ made up of tokens $t_{i}$. The set of all labels is denoted by $\mathcal{Y} = \{y_{i} | \forall i \in [i,|\mathcal{Y}|]\}$.

\paragraph{Saliency Representation}
\our employs salient tokens of a given document to identify which parts of the input should be attended to. We represent the set of all salient tokens of an input document as $\mathcal{E} = \{t_{i}| \forall i \in [1, |\mathcal{E}|]\}$ \cite{simonyan2014deep}, where token $t_{i}$ is salient.

The \our framework is depicted in Figure \ref{fig:main}, which incorporates both input text and saliency representations to learn contextualized mappings that are mapped to both text and saliency classifiers.

\subsection{Iterative Pseudo-Label Generation}

\paragraph{Pseudo-Text Classification Label Generation}
Using a PLM $\mathcal{T}^{C}$, we first derive pseudo-text classification labels automatically using only input text. For example, given the sentence "I really don't like The Green Bay packers" in Figure \ref{fig:main}, we feed this sentence through $\mathcal{T}^{C}$ to determine the appropriate classification label (in this case, negative sentiment). We formally define this query process using $\mathcal{D}$ as the input document to generate a pseudo-text classification label ${y}^{T}$ below:
\begin{equation} \label {eq:txtlabel}
    \hat{y}^{C} = \mathcal{T}^{C}(\mathcal{D}).    
\end{equation}

\paragraph{Pseudo-Explanation Label Generation}
It is possible that $\mathcal{T}^{C}$ may not produce confident predictions. For instance, $\mathcal{T}^{C}$ may classify the example sentence in Figure \ref{fig:main} as positive sentiment because of the words "really" and "like", disregarding the phrase "don't like". To further enhance these pseudo-text classification label predictions, we utilize another PLMs $\mathcal{T}^{E}$ that captures the reasoning of $\mathcal{T}^{C}$; namely, identifying the salient tokens in the input that were responsible for the pseudo-text classification label. 

Formally, for a given input document $\mathcal{D}$ and previously generated pseudo-text classification label $\hat{y}^{C}$, we query $\mathcal{T}^{E}$ to determine the salient tokens based on the predicted label:
\begin{equation} \label{eq:saliency}
    \hat{y}^{E} = \mathcal{T}^{E}(\mathcal{D}, \hat{y}^{C}),
\end{equation}
where $\hat{y}_{i}^{E}$ is a binary vector with cardinality $|\mathcal{D}|$ that's formulated based on the following equation: 
\begin{equation}
    \begin{cases}
        \mathcal{D}_{i} \text{\, is salient,} \,\, \,\, \,\, \,\, \,\, \,\, \, \hat{y}^{E}_{i} = 1\\
        \mathcal{D}_{i} \text{\, is not salient,} \,\,\,\,  \hat{y}^{E}_{i} = 0.
    \end{cases}
\end{equation}
The generation of pseudo-label text classification and explanation labels, respectively, form one \textit{round}.

\paragraph{Iterative Mutual Enhancement}
\label{sec:iterativenhance}
Using the pseudo-text classification and explanation labels generated, we once again query $\mathcal{T}^{C}$, but now we additionally provide the pseudo-explanation labels as input. For example, the sentence in round 1 of Figure \ref{fig:main} and the salient tokens (highlighted in red) are used as input to the classification prompt, which is fed into $\mathcal{T}^{C}$. This extension of equation \ref{eq:txtlabel} is defined below:
\begin{equation} \label{eq:txtenhanced}
    \hat{y}^{C} = \mathcal{T}^{C}(\mathcal{D}, \hat{y}^{E}). 
\end{equation}
We repeat equations \ref{eq:txtenhanced} and \ref{eq:saliency}, respectively, to ensure high confidence in both $\mathcal{T}^{C}$ and $\mathcal{T}^{E}$ predictions, i.e., the predictions from both PLMs do not further change after one round.

One important consideration for this first step of pseudo-label generation is the choice of PLMs. In our experiments, we use \textsc{FLAN-T5} \cite{flant5} as $\mathcal{T}^{C}$ for the text classification label generation, and \textsc{BERT} \cite{bert} as $\mathcal{T}^{E}$ for the explanation label generation. 

\subsection{Explainable Multi-Task Architecture}
Once $\mathcal{T}^{C}$ and $\mathcal{T}^{E}$ have generated confident labels, we then input both of these into a multi-task text classification model. In Figure \ref{fig:main} for example, we take the "negative" text classification label and the "really don't like" salient labels as input.

Specifically, we first tokenize the input document $\mathcal{D}$ using a BERT-based \cite{bert} tokenizer. We then pass this tokenized document into our BERT-based \cite{bert} multi-task model and extract the following information from the model:
\begin{equation}
    l^{C}, \mathbold{A} = \mathcal{T}(\mathcal{D}),
\end{equation}
where $l^{C}$ is the loss of the text classification task and $\mathbold{A} \in \mathbb{R}^{L \times H \times |\mathcal{D}| \times |\mathcal{D}|}$ is the multi-head attention tensor. $L$ is the number of layers, and $H$ is the number of attention heads in $\mathbold{A}$ from the BERT-based \cite{bert} model. We extract the attention matrix $\Tilde{\mathbold{A}} \in \mathbb{R}^{|\mathcal{D}| \times |\mathcal{D}|}$ from the last layer and the last attention head of $\mathbold{A}$. We then apply a linear classifier $\textbf{W} \in \mathbb{R}^{|\mathcal{D}| \times 1}$ to this attention matrix $\Tilde{\mathbold{A}}$:

\begin{equation} \label{eq:4}
\hat{y} = \Tilde{\mathbold{A}}\textbf{W} + \mathbold{b}
\end{equation}
where $\mathbold{b} \in \mathbb{R}^{|\mathcal{D}| \times 1}$ is the bias vector. We apply a sigmoid layer $\sigma(\cdot)$ on top of a binary cross-entropy loss function to get the attention-based loss $l^{E}$ of the saliency word prediction task:
\begin{equation}
\small
    l^{E} = -w[y \cdot log \sigma(\hat{y}) + (1 - y)
    \cdot log(1 - \sigma(\hat{y})],
\end{equation}

Our multi-task loss function is thus a linear combination of the aforementioned loss as well as the loss $l^{C}$ from the text classification task:

\begin{equation}
l = l^{C} + \lambda l^{E},
\end{equation}
where $\lambda \in [0,1]$ is a hyper-parameter controlling the performance balance between the text classification and saliency word prediction.

\section{Experiments}

\subsection{Experimental Setup}
\paragraph{Datasets}
We use six datasets below in our experiments. Details about the dataset statistics are shown in Table \ref{tab:dataset}.
\begin{itemize}[leftmargin=*]
\item \textbf{AGNews} \cite{DBLP:journals/corr/ZhangZL15} is a popular text classification dataset. It consists of news articles collected from the AG's online news corpus, with articles from four different categories: World, Sports, Business, and Science/Technology.

\item \textbf{20Newsgroup} \cite{LANG1995331} is another well-known benchmark dataset. The dataset consists of documents from 20 different newsgroups, covering a wide range of topics. 

\item \textbf{UCINews} \cite{misc_news_aggregator_359} collects a substantial number of news articles from March 10, 2014, to August 10, 2014, covering four categories: Entertainment, Technology, Business, and Health.

\item \textbf{IMDB} \cite{zaidan-etal-2007-using} contains movie reviews from IMDB, with each review accompanied by a sentiment label indicating whether the review is positive or negative.

\item \textbf{Twitter}\footnote{https://www.kaggle.com/competitions/tweet-sentiment-extraction} is a collection of tweets that have been labeled or annotated with sentiment labels, indicating whether the sentiment expressed in the tweet is positive, negative, or neutral. 

\item \textbf{MIMIC-III} \cite{johnson2018mimic} is a public electronic health records (EHRs) database with patient discharge summaries as text and diagnostic-related group (DRG) codes as class labels used in our experiments.

\end{itemize}

\begin{table}
  \centering
  \small
  \caption{Dataset statistics, depicting the sizes of the training, testing, and development set as well as the total number of classes.} 
  \begin{tabular}{l l l l l}
  \toprule
  \textbf{Datasets} & \textbf{\# Train} & \textbf{\# Dev} & \textbf{\# Test} & \textbf{\# Class} \\ 
\midrule[1pt]
  AGNews & 108,000 & 12,000 & 7,600 & 4 \\ 
  \midrule
  20Newsgroup & 14,609 & 1,825 & 1,825 & 6 
  \\
  \midrule
  UCINews & 26,008 & 2,560 & 27,556 & 4 \\ 
  \midrule
  IMDB & 16,000 & 200 & 200 & 2 \\
  \midrule 
  Twitter & 21,983 & 2,747 & 2,748 & 3 \\
  \midrule
  MIMIC-III & 20,266 & 2,252 & 2,252 & 369 \\
  \bottomrule
  \end{tabular}
  \label{tab:dataset}
\end{table}

\paragraph{Baselines}
Our baselines include both fully supervised and weakly supervised text classification methods below.
\begin{itemize}[leftmargin=*]
    \item \textbf{CNN} \cite{simonyan2014deep} is a fully supervised baseline that trains a Convolutional Neural Network model using the labeled data.
    \item \textbf{BERT} \cite{bert} is also a fully supervised baseline that trains a transformer model using the labeled data.

    \item \textbf{Clinical-BERT} \cite{alsentzer2019publicly} is a supervised baseline that trains a the BERT model on the clinical text.
    
    \item \textbf{Dataless} is a weakly supervised baseline that utilizes vector similarity to analyze the correlation between documents and labels and predicts based on the maximum cosine similarity.
    \item \textbf{WeSTClass} \cite{westclass} is a weakly supervised baseline that generates pseudo-labels and pseudo-samples to pre-train a neural network, followed by self-training.
    \item \textbf{LOTClass}\footnote{https://github.com/yumeng5/LOTClass} \cite{lotclass} is a weakly supervised baseline that utilizes PLMs to construct a keyword vocabulary for the pseudo-label generation.
    \item \textbf{ConWea}\footnote{https://github.com/dheeraj7596/ConWea} \cite{conwea} expands the keyword vocabulary based on contextual representations of the labels and the corpus.
    \item \textbf{XClass}\footnote{https://github.com/ZihanWangKi/XClass} \cite{xclass} uses the clustering method to choose the representative documents for each class.
    \item \textbf{WDDC-MLM}\footnote{https://github.com/HKUST-KnowComp/WDDC} \cite{wddc} employs a masked language model to generate signal words. It combines the generated words with category names and utilizes them for training.
    \item \textbf{WDDC-Doc} \cite{wddc} is the same as WDDC-MLM except that the supervision signals come from the document itself.
\end{itemize}

\paragraph{Evaluation Metrics} 
We use Micro-F1 and Macro-F1 as the evaluation metrics to compare the performance of the text classification methods. More details can be found in Appendix \ref{appendix:eval}.

\paragraph{Parameter Settings} 
For each baseline method, we use the default parameter settings as reported in the original papers. More details about the parameter settings of \our can be found in Appendix \ref{appendix:parameter}.

\subsection{Main Results}
\label{sec:mainResults}

\begin{table*}[t]
\caption{Micro- and Macro-F1 scores of baseline methods compared with \our. \our results are based on the optimal number of rounds associated with each dataset.}    
  \centering
    \begin{tabular}{c cc cc cc }
    \toprule
     \multirow{2}{*}{\textbf{Methods}} &
     \multicolumn{2}{c}{\textbf{AGNews}} & \multicolumn{2}{c}{\textbf{20Newsgroup}} &
     \multicolumn{2}{c}{\textbf{UCINews}} \\ 
     & \textbf{Micro} & \textbf{Macro} & \textbf{Micro} & \textbf{Macro} & \textbf{Micro} & \textbf{Macro} \\ 
    \midrule  
      CNN \cite{cnntxt}  &  0.9025 & 0.9025  & 0.9397 & 0.9310 &  0.9002 & 0.8998  \\
      BERT \cite{bert} &  \textbf{0.9305} & \textbf{0.9306} & \textbf{0.9660} & \textbf{0.9569} & \textbf{0.9313} & \textbf{0.9315}\\  
    \midrule
      Dataless \cite{dataless} &  0.6855 & 0.6844  &  0.5000 & 0.4700 &  0.6248 & 0.6253  \\
      WeSTClass \cite{westclass}  &  0.8279 & 0.8268  &  0.5300 & 0.4300  &  0.6983 & 0.6999 \\
      LOTClass \cite{lotclass} &  0.8659 & 0.8656 &  0.6121 & 0.5586  &  0.7320 & 0.7236 \\ 
      ConWea \cite{conwea} &  0.7443 & 0.7401 & 0.6200 & 0.5700 & 0.3293 & 0.3269\\
      X-Class \cite{xclass}  & 0.8574 & 0.8566 & 0.6515 & 0.6316 & 0.6885 & 0.6962 \\ 
      WDDC-MLM \cite{wddc} & \textbf{0.8826} & \textbf{0.8825} & 0.8121 & 0.6882 & \underline{0.8150} & \underline{0.8134} \\ 
      
      WDDC-Doc \cite{wddc} & 0.8668 & 0.8657 & \textbf{0.8570} & \textbf{0.8250} & 0.7814 & 0.7772 \\
      \midrule  
      \our & \underline{0.8820} & \underline{0.8815} & 0.7529 & \underline{0.7130} & \textbf{0.8395} & \textbf{0.8387} \\ 
        \bottomrule
    \end{tabular}%
  \label{tab:our result}%
\end{table*}

Our main results are displayed in Table \ref{tab:our result}. \our outperforms baselines in the UCINews while providing comparable results on AGNews. Additionally, we provide results for Twitter and IMDB datasets in Table \ref{tab:imdb_twitter}, as they have ground truth salient labels. It is our belief that the weakly-supervised SOTA on IMDB is in large part due to the sheer volume of sentiment in the dataset, as there are a total of 5618 ground truth salient sub-sequences \cite{diag}. It is interesting to note that the \our method does better than the BERT-based \cite{bert} model reported in \cite{diag}. Results in Table \ref{tab:mimic-iii} future demonstrate the domain generalizability of \our to the clinical text.

We hypothesize much of the performance dropoff in 20Newsgroup could be due to labels not being completely disjoint \cite{wddc}. For example, the "electronics" fine-grained class is categorized under the "science" class, although one could argue it would be more appropriate to classify instances of type "electronics" in the "computer" class \cite{LANG1995331}. 

\begin{table}[t]
\caption{F1 scores of BERT \cite{bert} baseline against \our variants. \textsc{XAI-Class-WS} is the weakly supervised \our variant, and \textsc{XAI-Class-FS} is the fully supervised \our. The backbone for all models is \textsc{BERT-base}.}
    \centering
\begin{tabular}{|c|c|c|}
    \toprule
    \multicolumn{1}{c}{\textbf{Model}} & \multicolumn{1}{c}{\textbf{Dev}} & \multicolumn{1}{c}{\textbf{Test}} \\
    \midrule

     \multicolumn{3}{c}{\textbf{Dataset: IMDB}} \\
     
     \multicolumn{1}{c}{BERT \cite{bert}} & \multicolumn{1}{c}{0.859} & \multicolumn{1}{c}{0.856 } \\

     \multicolumn{1}{c}{\textsc{XAI-Class-FS}} & \multicolumn{1}{c}{0.895} & \multicolumn{1}{c}{\textbf{0.878}} \\

     \multicolumn{1}{c}{\textsc{XAI-Class-WS}} & \multicolumn{1}{c}{\textbf{0.915}} & \multicolumn{1}{c}{0.864} \\
     
     \midrule 

      \multicolumn{3}{c}{\textbf{Dataset: Twitter}} \\
     
     \multicolumn{1}{c}{BERT \cite{bert}} & \multicolumn{1}{c}{0.772} & \multicolumn{1}{c}{0.781 } \\

     \multicolumn{1}{c}{\textsc{XAI-Class-FS}} & \multicolumn{1}{c}{\textbf{0.784}} & \multicolumn{1}{c}{\textbf{0.792}} \\

     \multicolumn{1}{c}{\textsc{XAI-Class-WS}} & \multicolumn{1}{c}{0.612} & \multicolumn{1}{c}{0.634} \\
     
     \bottomrule      
     
\end{tabular}
    \label{tab:imdb_twitter}
\end{table}

\begin{table}[t]
\caption{Macro- and Micro-F1 scores of Clinical-BERT \cite{alsentzer2019publicly} baseline against \our on the MIMIC-III dataset. }
    \centering
\begin{tabular}{|c|c|c|}
    \toprule
    \multicolumn{1}{c}{\textbf{Model}} & \multicolumn{1}{c}{\textbf{Micro}} & \multicolumn{1}{c}{\textbf{Macro}} \\
    \midrule
     \multicolumn{1}{c}{Clinical-BERT} & \multicolumn{1}{c}{0.256 } & \multicolumn{1}{c}{0.106} \\
     \multicolumn{1}{c}{\our} & \multicolumn{1}{c}{\textbf{0.292 }} & \multicolumn{1}{c}{\textbf{0.122} } \\
     \bottomrule     
\end{tabular}
    \label{tab:mimic-iii}
\end{table}

\subsection{Ablation Study}
\begin{figure*}[t]
     \centering
     \begin{subfigure}[b]{0.49\textwidth}
         \centering
         \includegraphics[width=\textwidth]{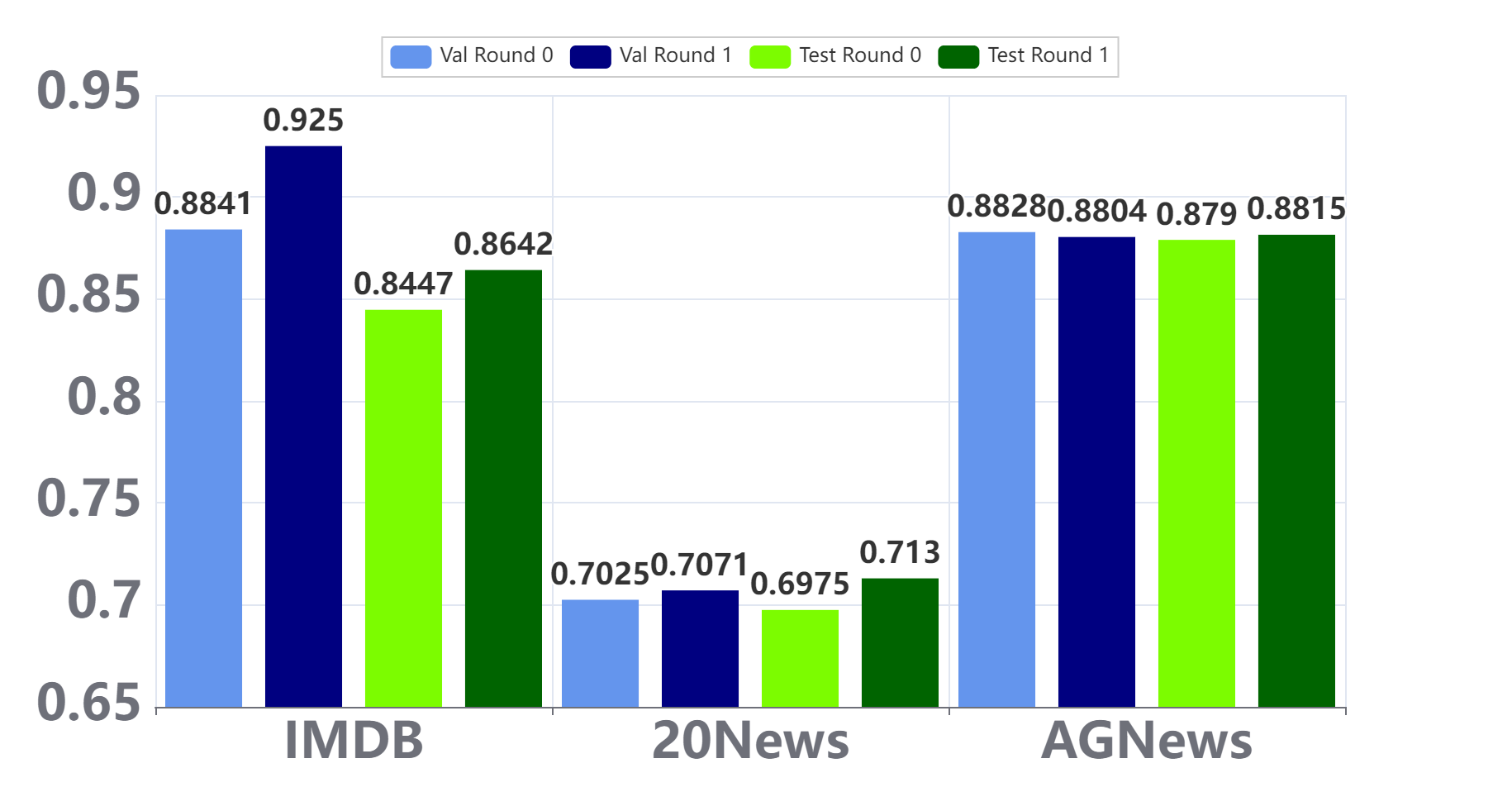}
         \caption{Macro-F1 Scores}
         \label{fig:f1macro}
     \end{subfigure}
     \hfill
     \begin{subfigure}[b]{0.49\textwidth}
         \centering
         \includegraphics[width=\textwidth]{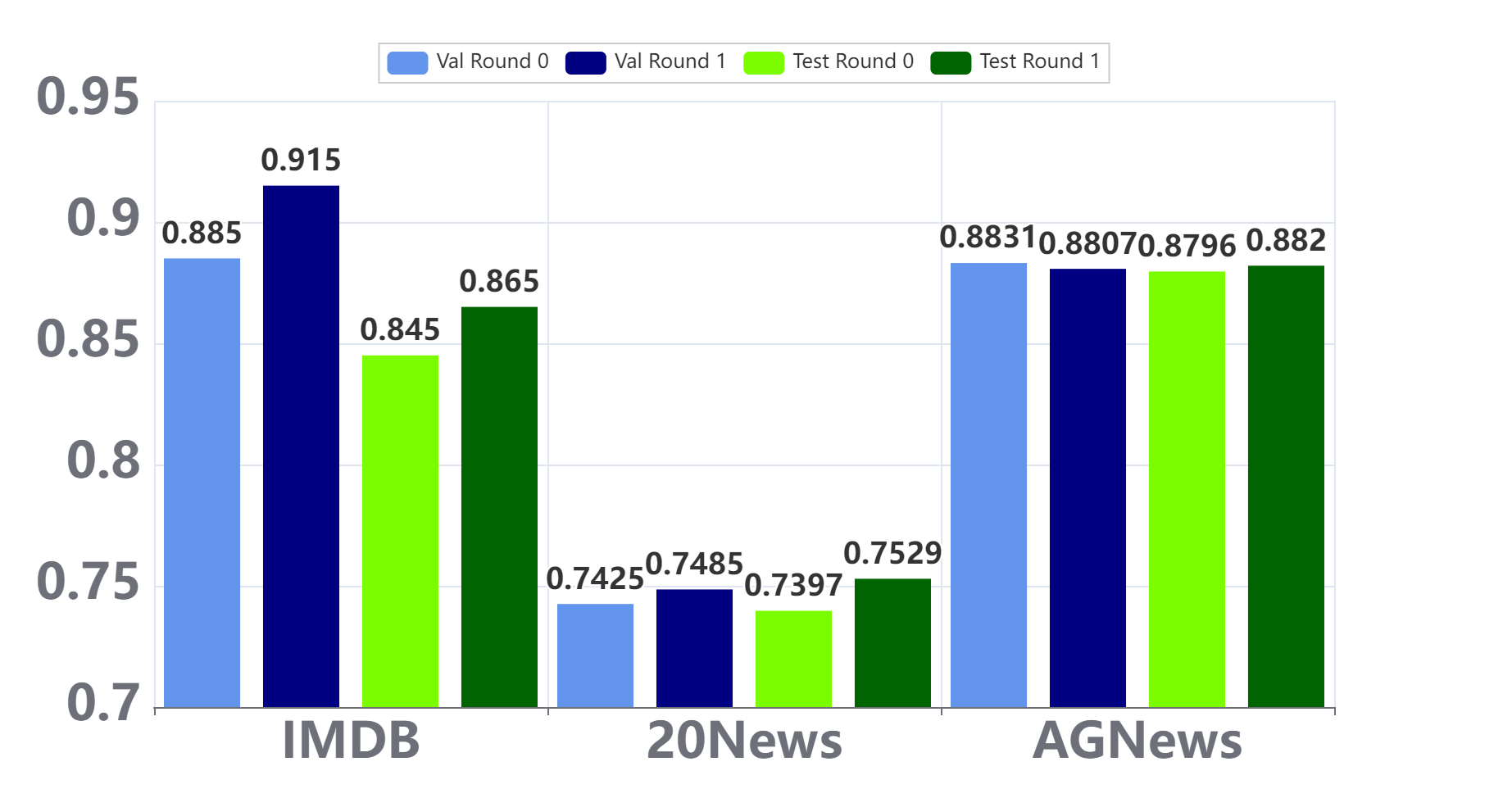}
         \caption{Micro-F1 Scores}
         \label{fig:f1micro}
     \end{subfigure}
     \caption{Test and validation Macro-F1 and Micro-F1 scores of two rounds of \our. The results are reported on both the development set and test set from three datasets: IMDB, 20News, and AGNews.}
     \vspace{-4mm}
    \label{fig:f1macro_f1micro}
\end{figure*}

To determine the effectiveness of iterative mutual enhancement \ref{sec:iterativenhance}, we identify the performance of datasets across multiple rounds. Figure \ref{fig:f1macro} and Figure \ref{fig:f1micro} show these results, clearly indicating that the performance increases when iterating up to a specified number of rounds. It should be noted that the optimal number of rounds is dependent on the dataset, with datasets that have high performance without many rounds most likely requiring fewer rounds than otherwise.

\subsection{Explainability Study}
\begin{figure*}[t]
     \centering
     \begin{subfigure}[b]{0.49\textwidth}
         \centering
         \includegraphics[width=\textwidth]{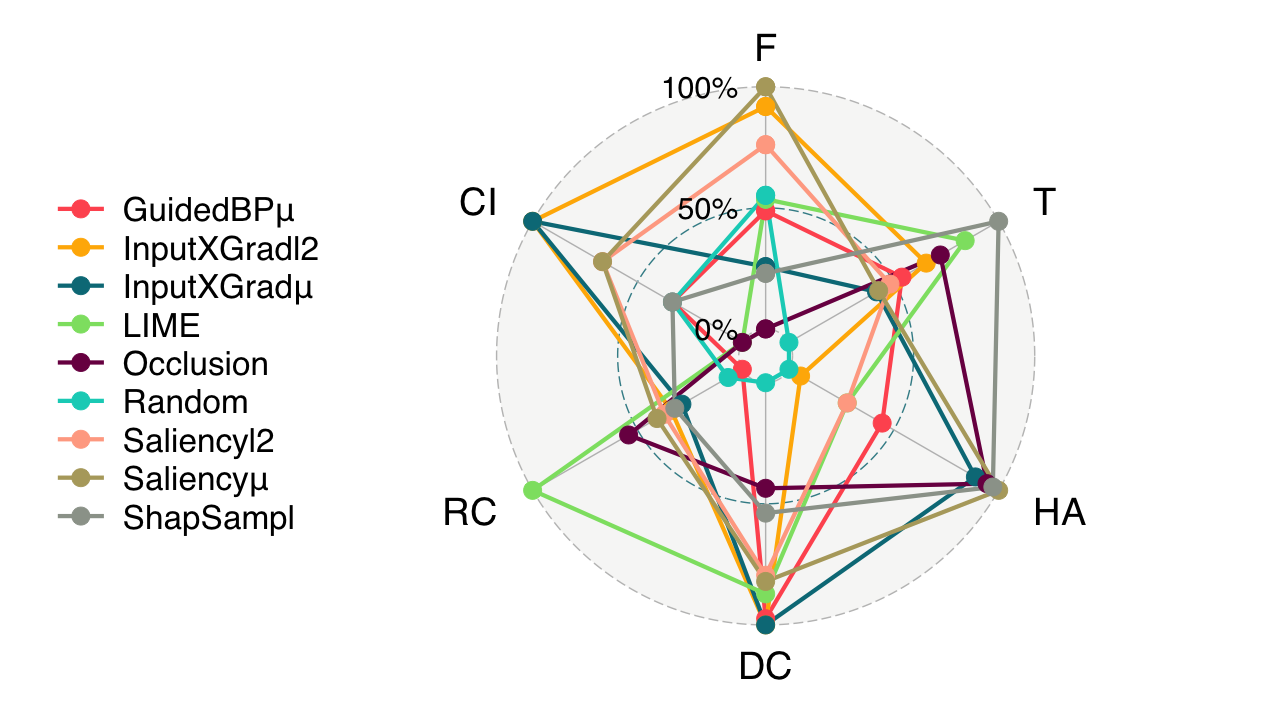}
         \caption{Clinical-BERT}
         \label{fig:clinical-bert}
     \end{subfigure}
     \hfill
     \begin{subfigure}[b]{0.49\textwidth}
         \centering
         \includegraphics[width=\textwidth]{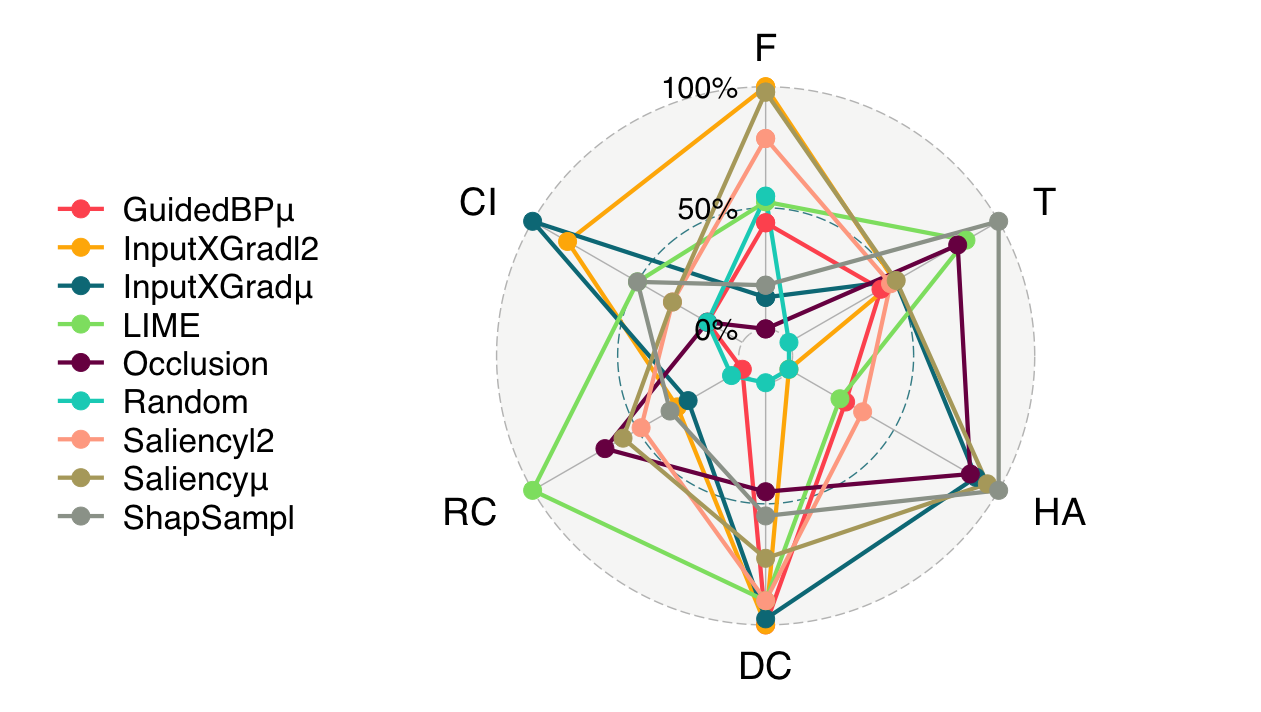}
         \caption{\our}
         \label{fig:xai-former}
     \end{subfigure}
     \caption{Explainability of Clinical-BERT and \our using six explanation techniques (Saliency, InputXGradient, Guided Backpropagation, Occlusion, Shapley Value Sampling, and LIME) on five explanation evaluation metrics (HA, CI, F, RC, DC) on the MIMIC-III dataset. A larger area in the hexagon indicates a better performance.}
     \vspace{-2mm}
    \label{fig:clinical-xai}
\end{figure*}

To evaluate the explainability of \our over the baseline methods, we qualitatively assess the expandability of Clinical-BERT and \our using six explanation techniques (Saliency \cite{simonyan2014deep}, InputXGradient \cite{inputx}, Guided Backpropagation \cite{guidedbp}, Occlusion \cite{occlusion}, Shapley Value Sampling \cite{shapely}, and LIME \cite{lime}) over five explanation evaluation metrics \cite{atanasova-etal-2020-diagnostic} (Agreement with Human Rationales (HA), Confidence Indication (CI), Faithfulness (F), Rationale Consistency (RC), and Dataset Consistency (DC)) on the MIMIC-III dataset. The results in Figure \ref{fig:clinical-xai} demonsrate that \our improved the model explainability by capturing the saliency information during the training process.
More results on the explainability case study can be found in Appendix \ref{appendix:exp_case}.

\begin{table*}[h!t]
\caption{Sample of instances with incorrect/ambiguous ground truths in the 20Newsgroup dataset.}
  \centering
  \small
  \begin{tabular}{m{7.7cm} m{1.3cm} m{1.9cm} m{3.2cm}}
  \toprule
  \textbf{Input Text} & \textbf{Prediction} & \textbf{Ground Truth} & \textbf{Predicted Salient Words} \\ \midrule[1pt]
  
  72 Chevelle SS for sale. I do not want to sell this car, but I need money for college. [...] 1972 chevelle super sport rebuilt 402 [...]  \$ 5995 or best offer. Call dennis at [...] & Sale & Sports & sale, money, sport \\ \midrule
  
  For the system, or ‘family’, key would appear to be cryptographically useless. [...] The same key is used for both encryption and decryption. & Computer & Science & cryptographically, encryption, key \\ \midrule

   What exactly is an IBM 486 SLC processor? Could someone please tell me if the 486 SLC and 486 SLC2 processors IBM is putting in their Thinkpad 700's. & Computer & Science & IBM, processor, 486 \\ \midrule

   Cultural enquiries more like those who use their backs instead of their minds [...] intolerant of anything outside of their group or level of understanding [..] there is no justification for taking away individuals freedom in the guise of public safety. & Politics  & Sports & cultural, freedom \\
   
    \bottomrule
  \end{tabular}  
  \vspace{-4mm}
  \label{tab:case_study}
\end{table*}

\subsection{Case Study}
We further explore some cases with incorrect/ambiguous ground truths for multiple reasons, depicted in Table \ref{tab:case_study}. The text in the first row of Table \ref{tab:case_study} is (most likely) supposed to be assigned to the "sale" class, but is instead labeled with the "sports" class as ground truth, most likely because the word "sport" appears in the text. \our predicted the "sale" class, even though it determined that "sport" was a salient token. This suggests that the model is robust to a small number of words dictating the classification prediction. The second row in Table \ref{tab:case_study} coincides with the cryptograph example in section \ref{sec:mainResults}, where one could argue all salient words picked up by the model could be categorized under the term "computer", instead of the ground truth "science". The last two rows of Table \ref{tab:case_study} appear to be  mislabelled, as the third row's text talks exclusively about processors and the fourth example talks only about political issues, yet they are labeled as "science" and "sports", respectively.

\section{Conclusion}
We propose \our, a novel explanation-enhanced extremely weakly-supervised text classification method that incorporates word saliency prediction as an auxiliary task. \our employs a multi-round question-answering process to generate pseudo-training data and trains a multi-task framework that simultaneously learns both text classification and word saliency prediction. Experiments demonstrate that \our has superior performance over baselines for both model performance and explainability. Future work includes extending \our to the multi-label setting.


\clearpage

\section*{Limitations}
\our, although effective, operates under the assumption of a disjoint label space and is not specifically tailored for fine-grained or multi-label text classification tasks. As a result, it may not perform optimally on datasets like 20Newsgroup, where there are instances where ground truth labels have some degree of overlap. However, exploring weakly-supervised methods for fine-grained, multi-label text classification is an intriguing direction for future research.
Furthermore, it's important to note that \our requires careful consideration when selecting the number of rounds of question answering. It is not designed to scale to a large number of rounds, and typically, no more than three rounds are used. This limitation arises because each round involves two queries for the question answering models: one for generating text classification labels and the other for saliency word generation. This process can be computationally expensive, necessitating a mindful balance between computational resources and desired performance.

\section*{Ethics Statement}
Given our current methodology, we do not anticipate any significant ethical concerns. We have utilized datasets and models from open-source domains, promoting transparency and accessibility of information. Text classification is a well-established task in natural language processing, widely studied and applied in various domains.
However, we acknowledge that our architecture relies on PLMs, which may make decisions based on biases present in the training data. Although our experiments have not revealed any apparent performance issues related to bias, it is important to recognize that this observation may be limited to the datasets we have used. It is crucial to remain vigilant and continue exploring ways to mitigate and address biases that may arise from the use of pre-trained models.


\clearpage
\appendix

\section{Evaluation Metrics}
\label{appendix:eval}


The evaluations we use, Micro-F1 and Macro-F1 as defined below:

\newcommand{\Fone}{F1}
\newcommand{\Fmacro}{F1_{\text{macro}}}
\newcommand{\Fmicro}{F1_{\text{micro}}}

\newcommand{\tp}{\text{TP}}
\newcommand{\fp}{\text{FP}}
\newcommand{\fn}{\text{FN}}

\newcommand{\FoneFormula}{
    \Fone = \frac{2 \cdot \tp}{2 \cdot \tp + \fp + \fn}
}

\newcommand{\FmacroFormula}{
    \Fmacro = \frac{1}{n} \sum_{i=1}^{n}{\Fone_i}
}

\newcommand{\FmicroFormula}{
    \Fmicro = \frac{2 \cdot \sum_{i=1}^{n}{\tp_i}}{2 \cdot \sum_{i=1}^{n}{\tp_i} + \sum_{i=1}^{n}{\fp_i} + \sum_{i=1}^{n}{\fn_i}}
}

\noindent
$$ \FoneFormula $$

\noindent
$$\FmicroFormula $$

\noindent
$$\FmacroFormula$$

\noindent
where TP is true positive, FP is false positive, and FN is false negative. We use the sklearn\footnote{https://scikit-learn.org/stable/} library to obtain these metrics.

\section{Parameter Settings}
\label{appendix:parameter}

\paragraph{Runtime Analysis}
We conduct all of our experiments on an NVIDIA DGX A100 GPU (640GB). The run times for optimal configurations across all datasets can be found in Table \ref{tab:run time}.

\begin{table}[h]
\caption{Average run time for each dataset for best hyper-parameter configuration.}
    \centering
    \begin{tabular}{c c}
        \toprule
         Dataset &  Runtime (hours) \\
        \midrule
         AGNews & 10 \\ 
         20NewsGroup & 4 \\ 
         UCINews & 4 \\ 
         IMDB & 1 \\ 
         Twitter & 3 \\
         \bottomrule
    \end{tabular}
    \label{tab:run time}
\end{table}

\begin{table*}
\caption{Optimal hyper-parameters for \our's results in Tables \ref{tab:our result} and \ref{tab:imdb_twitter}.}
    \centering
    \small
    \begin{tabular}{c c c c c c c c}
        \toprule
         Dataset & $\mathcal{T}^{C}$ & $\mathcal{T}^{E}$ & Round \# & $\lambda$ & Learning Rate & Dropout & \# Epochs \\ \midrule

         AGNews & \textsc{FLAN-T5-xxl} & \textsc{BERT-base} & 1 & 0.5 & $2e-05$ & 0.3 & 1 \\ \midrule

         20Newsgroup & \textsc{FLAN-T5-xl} & \textsc{BERT-base} & 2 & 0.7 & $2e-05$ & 0.3 & 3 \\ \midrule   
         
         UCINews & \textsc{FLAN-T5-xl} & \textsc{BERT-base} & 1 & 0.5 & $2e-05$ & 0.3 & 1 \\ \midrule      

         IMDB & \textsc{FLAN-T5-xl} & \textsc{BERT-base} & 1 & 0.9 & $2e-05$ & 0.4 & 3 \\ \midrule               

         Twitter & \textsc{FLAN-T5-xl} & \textsc{BERT-base} & 0 & 0.7 & $2e-05$ & 0.1 & 3 \\ \bottomrule
    \end{tabular}
    \label{tab:hyper_params}
\end{table*}

\paragraph{Hyper-parameters}
The optimal hyper-parameters for our results in Tables \ref{tab:our result} and \ref{tab:imdb_twitter} are listed in Table \ref{tab:hyper_params}. The possible values each of the hyper-parameters can take are listed below:
\begin{itemize}[leftmargin=*]
    \item $\mathcal{T}^{C}\in$ $\{$\textsc{FLAN-T5-small}, \textsc{FLAN-T5-base}, \textsc{FLAN-T5-large}, \textsc{FLAN-T5-xl}, \textsc{FLAN-T5-xxl}$\}$
    \begin{itemize}
        \item  PLM for psuedo-text classification label generation 
    \end{itemize}
    
    \item $\mathcal{T}^{E}\in$ $\{$\textsc{BERT-base}, \textsc{BERT-large}, \textsc{RoBERTA-base}, \textsc{RoBERTA-large} $\}$
    \begin{itemize}
        \item PLM for psuedo-saliency label generation
    \end{itemize}

    \item $\lambda \in \{0.5, 0.7, 0.9\}$
    \begin{itemize}
        \item Hyper-parameter for determining how much of the saliency loss should be incorporated
    \end{itemize}

    \item Round \# $\in \{0, 1, 2, 3\}$

    \item Learning Rate $\in$ \{$2e-04$, $2e-05$, $5e-05$\}

    \item Dropout $\in \{0.1, 0.2, 0.3, 0.4\}$

    \item \# Epochs $\in \{1,2,3\}$
\end{itemize}
We implement the PLMs in Python using the HuggingFace Transformer library\footnote{https://github.com/huggingface/transformers}.

\section{Explanability Case Study}
\label{appendix:exp_case}
\begin{figure*}[t]
     \centering
     \begin{subfigure}[b]{0.49\textwidth}
         \centering
         \includegraphics[width=\textwidth]{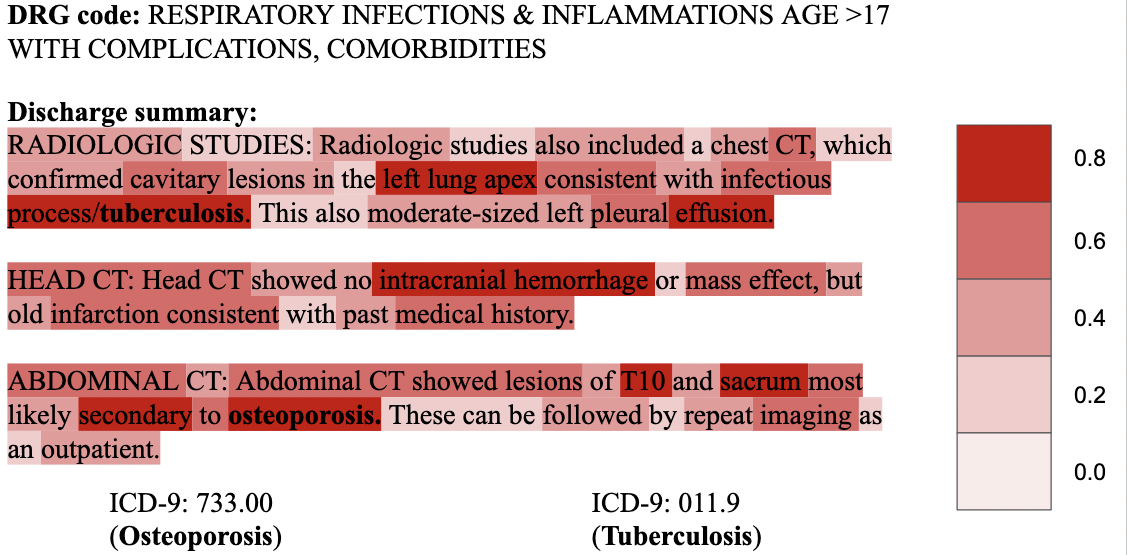}
         \caption{Clinical-BERT}
         \label{fig:clinical-bert}
     \end{subfigure}
     \hfill
     \begin{subfigure}[b]{0.49\textwidth}
         \centering
         \includegraphics[width=\textwidth]{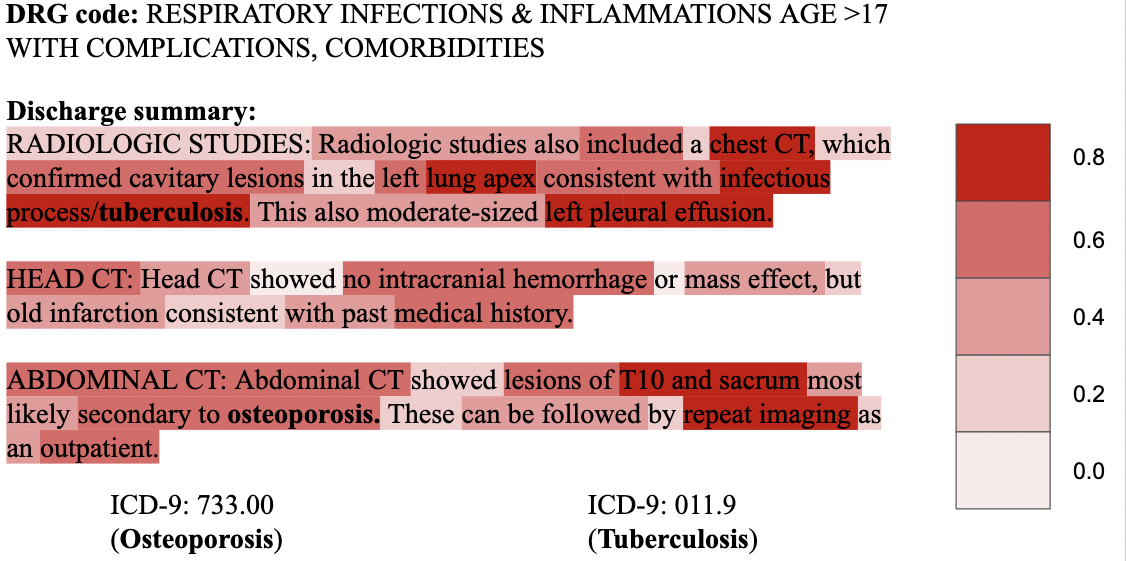}
         \caption{\our}
         \label{fig:xai-former}
     \end{subfigure}
     \caption{The attention distribution (heatmap) of of Clinical-BERT and \our. A darker red color indicates that the model assigns higher importance to that particular word for explaining the prediction of the DRG code.}
    \label{fig:exp_case}
\end{figure*}

To further evaluate the explainability of \our over the baseline methods, we qualitatively assess the explainability of Clinical-BERT and \our using attention distribution (heatmap). The results in Figure \ref{fig:exp_case} demonstrate that \our improved the model explainability by capturing the saliency information during the training process. The results align well with human-given ICD-9 codes as the explanation for the DRG code prediction.

\end{document}